%% file: PaperForCRV.tex
\documentclass[10pt,twocolumn,letterpaper]{article}

\usepackage[pagenumbers]{wacv} 

\usepackage{graphicx}
\usepackage{amsmath}
\usepackage{amssymb}
\usepackage{booktabs}
\usepackage{multirow} 
\usepackage{float}
\usepackage{tcolorbox}
\usepackage[pagebackref,breaklinks,colorlinks]{hyperref}

\newcommand{\diego}[2]{\textcolor[RGB]{80, 200, 120}{#2}}

\usepackage[capitalize]{cleveref}
\crefname{section}{Sec.}{Secs.}
\Crefname{section}{Section}{Sections}
\Crefname{table}{Table}{Tables}
\crefname{table}{Tab.}{Tabs.}

\def\wacvPaperID{1502} 
\def\confName{WACV}
\def\confYear{2025}

\begin{document}

\title{SenCLIP: Enhancing zero-shot land-use mapping for Sentinel-2 with ground-level prompting}
\author{Pallavi Jain\textsuperscript{1, 2, 6}, Dino Ienco\textsuperscript{2, 3, 5, 6}, Roberto Interdonato\textsuperscript{2, 4, 5, 6}, \\ Tristan Berchoux\textsuperscript{1} and Diego Marcos\textsuperscript{2, 6}\\
\textsuperscript{1}Mediterranean Agronomic Institute of Montpellier - CIHEAM-IAMM,
\textsuperscript{2} Inria,  \textsuperscript{3} INRAE, \\ \textsuperscript{4} Cirad, \textsuperscript{5} UMR TETIS, \textsuperscript{6} Univ. of Montpellier, Montpellier France \\
{\tt\small \{pallavi.jain, dino.ienco, roberto.interdonato, diego.marcos\}@inria.fr }\\
{\tt\small  berchoux@iamm.fr}
}
\maketitle

\begin{abstract}
Pre-trained vision-language models (VLMs), such as CLIP, demonstrate impressive zero-shot classification capabilities with free-form prompts and even show some generalization in specialized domains. However, their performance on satellite imagery is limited due to the underrepresentation of such data in their training sets, which predominantly consist of ground-level images. Existing prompting techniques for satellite imagery are often restricted to generic phrases like \textit{“a satellite image of…”}, limiting their effectiveness for zero-shot land-use/land-cover (LULC) mapping. To address these challenges, we introduce SenCLIP, which transfers CLIP’s representation to Sentinel-2 imagery by leveraging a large dataset of Sentinel-2 images paired with geotagged ground-level photos from across Europe. We evaluate SenCLIP alongside other state-of-the-art remote sensing VLMs on zero-shot LULC mapping tasks using the EuroSAT and BigEarthNet datasets with both aerial and ground-level prompting styles. Our approach, which aligns ground-level representations with satellite imagery, demonstrates significant improvements in classification accuracy across both prompt styles, opening new possibilities for applying free-form textual descriptions in zero-shot LULC mapping. Code, dataset and pretrained models are available at \url{https://github.com/pallavijain-pj/SenCLIP}
\end{abstract}

\section{Introduction}
\label{sec:intro}
Monitoring land-use and land-cover (LULC) is essential for understanding human impacts on the environment and assessing related risks~\cite{turner2007emergence}. Spaceborne sensors have long been used to map LULC, providing key data for policy-relevant indicators, especially in rural areas~\cite{anderson1976land}. These insights are crucial for sustainable development, land-use planning, habitat preservation, and effective natural resource management~\cite{hargreaves2021eo4rural,giuliani2020landdegradation}.
During the last decade, the evolution of deep learning has enhanced the reliability of LULC prediction based on remote sensing~\cite{zhu2017deep}. However, these approaches generally depend on large datasets for training the models via supervised learning, requiring a large initial effort and restricting them to a closed set of initial LULC classes.
In general, the class labels influence the learning process by guiding the optimisation of the network's parameters towards minimising the classification error for the known classes~\cite{he2016deep, krizhevsky2012imagenet}. Consequently, the network may prioritise learning features that are highly discriminative for these specific set of classes.
This limits the usefulness of the representation in recognising unseen classes without additional training or fine-tuning steps.

Recent advancements in vision-language models (VLMs) have revolutionised this paradigm. VLMs leverage Web-scale image/caption datasets to learn a representation that allows for zero-shot predictions across various tasks. 
These models operate by learning a joint semantic space for image/text pairs through contrastive learning strategies. Prominent examples of VLMs include CLIP~\cite{radford2021learning}, ALIGN~\cite{jia2021scaling}, and BLIP~\cite{li2023blip, li2022blip}, which aim to map both textual descriptions and visual representations into the same latent space, enabling direct semantic comparisons between the two modalities. 
Despite the significant strides made by VLMs, there are persistent challenges in the practical implementation of zero-shot approaches based on them. 
One of the obstacles lies in the manual prompting process intrinsic to these approaches, where the choice of vocabulary and specific textual cues for each class play a pivotal role. Even slight variations in prompt formulation, such as the inclusion or exclusion of articles like \textit{``a"} can wield considerable influence over the model's accuracy~\cite{zhou2022learning}. 
Moreover, achieving optimal performance often requires incorporating task-relevant information into the prompts. In specialised domains like remote sensing, specifying details such as \textit{``centered satellite photo''} or describing specific attributes like \textit{``photo of \{class\}: a type of broadleaf forest"} alongside the class name becomes imperative for accurate classification~\cite{radford2021learning, zhou2022learning, allingham2023simple}. 
This nuanced prompt construction ensures that the model captures essential contextual cues, leading to improved classification accuracy~\cite{zhang2024concept}, even if the added context is meaningless~\cite{roth2023waffling}.
The sensitivity of prompt engineering and prompt learning~\cite{zhou2022conditional} to a specific context underscores the critical role of prompting for harnessing the full potential of VLMs. 

Another significant challenge arises when attempting to move into specialised domains that are underrepresented in the training sets of VLMs, as is the case in remote sensing tasks. This results in challenges both for the image representation, which suffers due to the differences \diego{from}{to} the more common image modalities in terms of resolution, perspective and radiometry, and the text representation.
Indeed, the CLIP representation of satellite images tends to be aligned with coarse concepts (e.g. \textit{this is a satellite image}) that are of little help for understanding LULC~\cite{dhakal2023sat2cap}.
One potential solution lies in the development of large-scale VLMs specifically tailored for remote sensing applications. However, a significant obstacle arises due to the scarcity of textual descriptions associated with remote sensing data. Unlike other domains with abundant image/text pairs, remote sensing lacks a substantial corpus of annotated satellite imagery with corresponding textual descriptions.

Recent advancements like RemoteCLIP~\cite{liu2023remoteclip}, SkyCLIP~\cite{wang2024skyscript}, and GeoRSCLIP~\cite{zhang2023rs5m} have successfully integrated remote sensing data with captions and class labels to enhance VLMs. These methods use curated satellite imagery, improving the models' ability to understand spatial contexts and semantic relationships through supervised pre-training, leading to better performance.
In contrast, label-free approaches such as Sat2Cap~\cite{dhakal2023sat2cap} focus on cross-view learning by integrating geotagged ground-level images with high-resolution satellite imagery. This enables the model to associate local scene details with satellite observations, allowing it to interpret ground-level prompts more effectively. By transferring detailed ground-level knowledge to broader satellite contexts, these approaches improve land-use and land-cover classification accuracy.

In this work, we extend the Sat2Cap approach by leveraging the LUCAS dataset, which contains nearly one million geotagged images across the European Union, alongside Sentinel-2 medium-resolution imagery. Our key contributions are:

    \noindent\textbf{1) Alignment of Sentinel-2 representations with ground-level images.} We advance zero-shot LULC classification by directly aligning Sentinel-2 imagery with co-located ground-level images from LUCAS, which allows the model to understand the satellite imagery via textual descriptions associated with ground-level concepts, as done in Sat2Cap~\cite{dhakal2023sat2cap}. This is done without relying on labels or captions, in contrast to competing approaches~\cite{liu2023remoteclip,wang2024skyscript,zhang2023rs5m}. Unlike in Sat2Cap~\cite{dhakal2023sat2cap}, which focuses on higher-resolution satellite images, our approach bridges the visual representation gap between ground-level perspectives and Sentinel-2's medium-resolution (10 m) multi-spectral data. This is particularly valuable given Sentinel-2’s widespread use in environmental and agricultural applications, where capturing fine-grained details remains a challenge. 
    
    \noindent\textbf{2) New state-of-the-art in zero-shot LULC classification.} By coupling our SenCLIP model with a rich, LLM-generated, set of class-specific prompts, we obtain substantial improvements in terms of zero-shot LULC classification performance on both EuroSAT and BigEarthNet, validating the usefulness of the approach for real-world zero-shot LULC mapping. Additionally, we introduce a \textbf{simple prompt selection method}, aiming to optimise performance by identifying the most representative prompts for each class to enhance classification performance.


\section{Related work}
\label{sec:related_work}

\noindent \textbf{Zero-shot land use and land cover mapping:} The large-scale study of LULC with spaceborne sensors~\cite{rogan2004remote} has taken large strides in terms of performance thanks to the application of deep learning-based methods~\cite{zhu2017deep}.
Although determining some aspects of land cover, such as detecting water, evergreen forests, or built-up areas, is often done at an operational level at large scales~\cite{hansen2013high,buttner2014corine}, many aspects of land use are hard to solve employing a satellite perspective only and require the use of ground-level information~\cite{srivastava2020fine}.
This is even more so the case in a zero-shot setting, when no class labels are available during training~\cite{wu2023mixed}.
Indeed, the majority of previous work on zero-shot classification in remote sensing focuses on aerial or very high resolution ($<$1 m) satellite imagery, where minute details can be interpreted by humans or pretrained computer vision models~\cite{sumbul2017fine,li2017zero,li2021robust,li2023rs}, with most methods for medium resolution satellite imagery ($\approx 10$ m) focusing on the more relaxed few-shot setting, where a few training samples are available~\cite{russwurm2020meta,russwurm2022humans}.
In this paper, we leverage the fine-grained, easily detectable, details in geotagged ground-level images to enable zero-shot LULC classification using medium resolution satellite imagery.

\noindent \textbf{Prompt  tuning:} 
Manual prompt engineering for zero-shot tasks with VLMs often prioritises linguistic nuances over visual cues, potentially limiting accuracy. To address this, large language models (LLMs) are increasingly used to improve prompting methods for better accuracy and robustness~\cite{pratt2023does,mirza2023lafter}. Context Optimisation (CoOp) methods~\cite{zhou2022learning,zhou2022conditional} take a different route by learning non-textual prompts from training data, although they struggle with fine-grained tasks~\cite{zhang2024concept}. Recent approaches, like incorporating visual cues into prompts~\cite{zhang2024concept}, or adding noise to enhance robustness~\cite{roth2023waffling}, have also improved VLM performance. In our work, we adapt standard, LLM-based, and ground-level perspective prompting for remote sensing and close the visual representation gap through cross-view learning with geotagged images.

\noindent \textbf{Cross-view learning:} 
Geo-localised ground-level photos offer a promising avenue to leverage the descriptive capabilities of ground-level vision models, often surpassing those using remote sensing data. Cross-view methodologies provide insights into image similarity, localization, and orientation~\cite{lin2015learning,shi2020looking,shi2022beyond}. VIGOR~\cite{zhu2021vigor} employs contrastive learning to compare features between aerial- and street-view images, enhancing scene analysis from diverse viewpoints. Similarly, TransGeo~\cite{zhu2022transgeo} uses attention-guided non-uniform cropping to enrich aerial-view features with ground-level details. However, bridging the gap between ground-level and satellite features remains challenging, particularly with low-resolution data.
To address this, Sat2Cap~\cite{dhakal2023sat2cap} introduces a cross-view modeling framework that predicts CLIP embeddings for ground-level scenes using overhead imagery.  Sat2Cap focuses on retrieval tasks and does not address the challenges of LULC classification or the handling of lower-resolution satellite images. Our work extends the Sat2Cap approach by targeting medium-resolution Sentinel-2 imagery and rural LULC tasks, achieving higher precision. While Sat2Cap captures satellite image features effectively, we enhance this by integrating four directional ground-level images per location from the LUCAS dataset. This cross-view integration enriches semantic context, leading to more accurate Sentinel-2 image representations and a better understanding of both ground and overhead modalities.
\section{Method}
\label{sec:prop_approach}
\begin{figure*}[ht]
    \centering
        \centering
        \includegraphics[width=\textwidth]{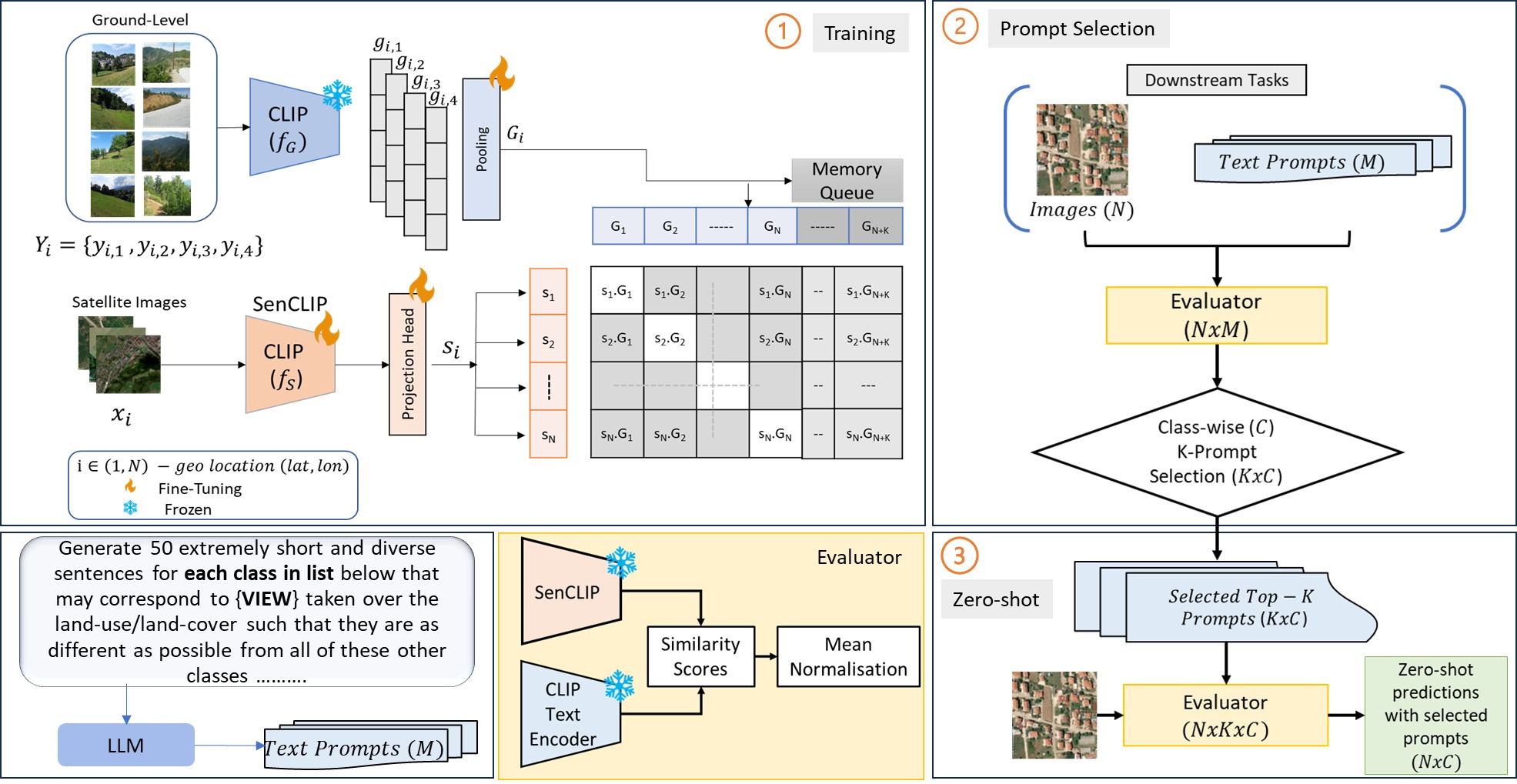}
        \caption{Architecture: The figure illustrates the three-step pipeline consisting of Pre-Training, Prompt Selection, and Zero-shot Predictions. It also demonstrates the prompt generation process from LLMs, which is utilised for prompt selection and then selected prompts for zero-shot prediction. }
        \label{fig:model_arch}
    \end{figure*}

Our method makes use of the rich and text-aligned representation provided by CLIP~\cite{radford2021learning} on ground-level images and transfers them to a satellite image representation via geotagged photos, similarly to~\cite{dhakal2023sat2cap}.
As shown in Fig. \ref{fig:model_arch}, we use two separate CLIP image encoders: one is kept frozen and provides frozen ground-level embeddings, while the other is fine-tuned for satellite data. The ground-level embeddings serve as targets, guiding the satellite image embedding to learn the semantic space that enables the alignment of satellite-derived data with the manifold of original CLIP embeddings from the ground-level perspective. 

\subsection{Self-supervised training dataset}

The ground-level images used in this study were obtained from the LUCAS dataset~\cite{d2020harmonised}, a comprehensive rural survey dataset providing Land Use and Land Cover information across Europe. The 2018 LUCAS survey consists of approximately 235,000 geotagged locations. Each location is associated with four directional images (taken from the north, east, west, and south), resulting in a total of around 900,000 images. This large-scale dataset offers a rich source of information for analyzing land use and cover patterns.

Using the LUCAS geolocations, we accessed Sentinel-2 data via the Planetary Computer API~\cite{microsoft_open_source_2022_7261897}. Our data retrieval focused on acquiring imagery from specific months and years corresponding to the LUCAS dataset. To ensure data quality, we applied a cloud coverage filter, selecting images with less than 10-20\% cloud cover. The obtained Sentinel-2 data included RGB bands with a resolution of 10 meters per pixel and scene dimensions of $100 \times 100$ pixels. More details on the dataset and its distribution across Europe are provided in supplementary. 

\subsection{Ground-level representation of satellite images}
For the two CLIP image encoders (frozen and trainable), pairs of ground-level images ($\mathbf{Y}$) and geo-located satellite images ($\mathbf{x}$) are utilised, denoted as $\{(\mathbf{Y}_1,\mathbf{x}_1),(\mathbf{Y}_2,\mathbf{x}_2),\dots,(\mathbf{Y}_N,\mathbf{x}_N)\}$. Here, $\mathbf{Y}_i = \{\mathbf{y}_{i,1}, \mathbf{y}_{i,2},\dots,\mathbf{y}_{i,K}\}$ represents the set of ground-level images corresponding to the $i^{th}$ location.

The frozen embeddings are obtained from ground-level images using the pre-trained CLIP encoder, denoted as $f_G$, such that ground-level image embeddings are computed as: $\mathbf{g}_{i,k}=f_G(\mathbf{y}_{i,k})$. Simultaneously, the satellite image encoder $f_{S}$ is initialised with the original CLIP image encoder and undergoes fine-tuning with Sentinel data. This results in our fine-tuned models, which we refer to as SenCLIP when combined with the projection head defined below.

\noindent\textbf{Pooling}
For each location, the frozen embeddings correspond to a set of ground-level images. To consolidate these into a single embedding $\mathbf{G}_i$ per location, represented by a set of quadruplet embeddings $\{\mathbf{g}_{i,1}, \mathbf{g}_{i,2},\dots,\mathbf{g}_{i,K}\}$, we explore two different pooling methods: average and attention pooling. Average pooling (AvgPool) is a simple and efficient approach that assigns equal importance to all four directional images. The embedding $\mathbf{G}_i$ is defined as follows:
\begin{equation}
\mathbf{G}_i = \frac{1}{K} \sum_{k=1}^{K} (\mathbf{g}_{i,k}).
\end{equation}

Attention pooling (AttPool) allows the model to focus on the most informative features from each location. The embedding $\mathbf{G}_i$ is obtained as follows: 
\begin{equation}
\mathbf{G}_i = \sum_{k=1}^{K} (w_{i,k} \cdot \mathbf{g}_{i,k}),
\end{equation}
where $w_{i,k}(\mathbf{g}_{i,k})$ represents the trainable attention weight for the $i$-th location and the $k$-th image, parameterised using a fully connected neural network.

\noindent\textbf{Projection head}
The projection head $H$, adapted from the implementation in~\cite{Shariatnia_Simple_CLIP_2021}, transforms the embeddings into a new space to capture richer relationships. It consists of a two-layer linear neural network, with GELU activation, dropout regularisation, and layer normalisation.
 
A residual connection is created by adding the output of the first linear layer to the output of the dropout layer.
The projection head is trained alongside the rest of the model. When combined with the trainable CLIP image encoder, the final model produces an embedding $\mathbf{s}$ for the satellite image $\mathbf{x}$, as shown below:
\begin{equation}
    \mathbf{s_i}=\text{SenCLIP}(\mathbf{x_i})=H(f_S(\mathbf{\mathbf{x_i}})).
\end{equation}

 \noindent\textbf{Training}
During training, two components are trained: the image encoder and a projection head for satellite images. Additionally, a pooling head is incorporated into the frozen encoder to consolidate ground-level quadruplets. The training process uses a dictionary implemented as a queue, inspired by the MoCo framework~\cite{he2020momentum}, to manage pooled frozen ground-level embeddings. This mechanism allows the reuse of encoded keys from previous iterations, with samples gradually replaced and the oldest batch removed to maintain consistency with newer samples. The training objective is to optimize the InfoNCE (Information Noise Contrastive Estimation)~\cite{oord2018representation} contrastive loss function between the pooled embeddings $\mathbf{G}_i$ from ground-level quadruplets and the fine-tuned satellite embeddings $\mathbf{s}_i$. This loss function evaluates both the similarity and dissimilarity between these two sets of embeddings, offering valuable insights into the effectiveness of the model's training process.
\begin{equation}
    \mathcal{L}_\text{InfoNCE} =  -\frac{1}{N} \sum_{i=1}^{N} \log \left ( \frac{\exp(\mathbf{G}_i \cdot \mathbf{s}_i/ \tau) }{ \sum_{j=1}^{N} \exp(\mathbf{G}_i \cdot \mathbf{s}_j/ \tau))} \right )
\end{equation}
where $\tau$ is the temperature and $N$, the number of samples.

\subsection{Prompting and zero-shot inference}

\noindent\textbf{Class-specific view-dependent prompts.}
We used an LLM to generate view-specific prompts encompassing aerial/overhead and ground-level views, as depicted in Fig.\ref{fig:model_arch}. Examples of generated prompts shown in supplementary. 
Based on the training described above, we expect SenCLIP to perform well on prompts describing LULC classes from a ground-level perspective. We generated a fixed number of prompts, $T$, for each class $c \in [1, \dots, C]$.

\noindent\textbf{Prompt-based zero-shot classification.}
The generated prompts, represented as vectors $\mathbf{a}_{c,t}$, each prompt can be considered to be an attribute of its corresponding class $c$. The similarity between a satellite image embedding $\mathbf{s}_i$ and the full set of text embeddings is determined through a dot product. 
In order to obtain the class scores, we employ Direct Attribute Prediction (DAP)~\cite{lampert2009learning}. The similarity between $\mathbf{s_i}$ and each attribute vector $\mathbf{a}_{c,t}$, calculated as $\mathbf{s}_i \cdot \mathbf{a}_{c,t}$, serves as a proxy for $p(\mathbf{a}_{c,t}|\mathbf{s}_i)$ i.e. probability of class attribute given the satellite image embedding.
The final classification assignment is then computed as
\begin{equation}
c_i = \arg\max_c \prod_{t=1}^T \frac{p(\mathbf{a}_{c,t}|\mathbf{s}_i)}{p(\mathbf{a}_{c,t})},
\end{equation}
where $p(\mathbf{a}_{c,t})$ is empirically estimated as the mean similarity of $\mathbf{a}_{c,t}$ with the full image set.

\noindent\textbf{Prompt selection method.}
This process involves computing a goodness score for each prompt $\mathbf{a}_{c,t}$, based on its similarity to the rest of the prompts. These scores are then used to identify the top prompts for each class.

Specifically, we use a weighted mean score to measure how well each prompt represents its corresponding class. Prompts that are more representative of their class characteristics are assigned higher weighted scores. This is achieved by calculating the mean within class similarity 
\begin{equation}
   \alpha_{c,t} =  \frac{\sum_{q=1}^T \mathbf{a}_{c,t} \cdot \mathbf{a}_{c,q}}{T}
\end{equation}
 for each class, and comparing it to the overall mean similarity scores
 \begin{equation}
   \beta_{c,t} =  \frac{\sum_{d=1}^C\sum_{q=1}^T \mathbf{a}_{c,t} \cdot \mathbf{a}_{d,q}}{C\cdot T}.
\end{equation}

 The ratio $w_{c,t} = \frac{\alpha_{c,t}}{\beta_{c,t}}$ effectively evaluates the representativeness of each prompt, emphasizing those with higher relevance to specific class characteristics. Higher weighted scores indicate prompts that more effectively encapsulate the essence of their class, serving as strong indicators of class-specific attributes. This approach aligns with the objective of prioritizing prompts that are more indicative and representative of their classes, while diminishing the influence of less relevant prompts.

\section{Experiments and results}
\label{sec:exp}
This section outlines the training implementation and hyperparameters, followed by the quantitative and qualitative results for SenCLIP. Quantitative results cover zero-shot inference, the impact of prompt selection, and performance improvements with LaFTer~\cite{mirza2023lafter}. Qualitative analysis using image captioning and cross-view image retrieval highlights the quality of the learned representations.

\subsection{Implementation details}
We fine-tuned SenCLIP using two backbone CLIP encoders: ResNet50 (RN50) and ViT-B/32. The fine-tuning strategy varied based on the model architecture: for RN50, all layers were fine-tuned, whereas for ViT-B/32, we fine-tuned only the last transformer block, linear layer, and the projection head. The AdamW optimizer, proposed by~\cite{loshchilov2017decoupled}, was used with initial learning rates (LR) of $10^{-5}$ for RN50 and $10^{-4}$ for ViT-B/32. Training was conducted over 20 epochs with a batch size of 32, incorporating a step scheduler with a step size of 5 and a decay multiplier of 0.95. The temperature parameter $\tau$ was set to 0.07 to scale the similarity scores. Data augmentation techniques included resizing, center cropping to $32\times 32$, random flipping, and rotation. For generating the prompts, we used GPT-3.5~\cite{ouyang2022training} as LLM. The models were trained on a single NVIDIA Titan X GPU.

\subsection{Quantitative results}
To evaluate the effectiveness of the learned representations, we utilised two well-established Sentinel-2 benchmark datasets: EuroSAT~\cite{helber2019eurosat} and BigEarthNet~\cite{sumbul2020bigearthnet}. EuroSAT contains images with 10 distinct, single-class land use/cover categories. On the other hand, BigEarthNet offers a more extensive set of annotations, with 19 multi-label classes. Consequently, Top-1 accuracy was used as the evaluation metric for EuroSAT, while mean average precision (mAP) was employed for BigEarthNet. All evaluations were performed in a zero-shot setting, where the models were not exposed to the training sets of the benchmarks. Classification was carried out by comparing image features with class-associated text features. 

\begin{table}[tbp]
\centering
\resizebox{\columnwidth}{!}{%
\begin{tabular}{clccc}
\hline
\multicolumn{1}{l|}{Model Arch}                & \multicolumn{1}{l|}{Prompt Templates/Models} & Generic        & Aerial         & Ground         \\ \hline
\multicolumn{5}{c}{EuroSAT}                                                                                                                      \\ \hline
\multicolumn{1}{c|}{\multirow{4}{*}{RN50}}     & \multicolumn{1}{l|}{CLIP}                    & 40.55          & 47.64          & 32.28          \\
\multicolumn{1}{c|}{}                          & \multicolumn{1}{l|}{RemoteCLIP}              & 25.20          & 29.95          & 22.13          \\
\multicolumn{1}{c|}{}                          & \multicolumn{1}{l|}{SenCLIP-AvgPool}         & 53.89          & 57.54          & 56.71          \\
\multicolumn{1}{c|}{}                          & \multicolumn{1}{l|}{SenCLIP-AttPool} & \textbf{56.53} & \textbf{57.78} & \textbf{57.95} \\ \hline
\multicolumn{1}{c|}{\multirow{6}{*}{ViT-B/32}} & \multicolumn{1}{l|}{CLIP}                    & 47.26          & 54.87          & 51.66          \\
\multicolumn{1}{c|}{}                          & \multicolumn{1}{l|}{RemoteCLIP}              & 44.74          & 48.95          & 43.21          \\
\multicolumn{1}{c|}{}                          & \multicolumn{1}{l|}{SkyCLIP50}               & 55.66          & 66.04          & 59.98          \\
\multicolumn{1}{c|}{}                          & \multicolumn{1}{l|}{GeoRSCLIP}               & \textbf{63.40} & 68.02          & 65.82          \\
\multicolumn{1}{c|}{}                          & \multicolumn{1}{l|}{SenCLIP-AvgPool}         & 61.18          & \textbf{71.22} & 65.54          \\
\multicolumn{1}{c|}{}                          & \multicolumn{1}{l|}{SenCLIP-AttPool} & 62.24          & 70.78          & \textbf{66.91} \\ \hline
\multicolumn{5}{c}{BigEarthNet}                                                                                                                  \\ \hline
\multicolumn{1}{c|}{\multirow{4}{*}{RN50}}     & \multicolumn{1}{l|}{CLIP}                    & 27.71          & 29.77          & 24.09          \\
\multicolumn{1}{c|}{}                          & \multicolumn{1}{l|}{RemoteCLIP}              & 23.04          & 33.00          & 20.23          \\
\multicolumn{1}{c|}{}                          & \multicolumn{1}{l|}{SenCLIP-AvgPool}         & 32.74          & 32.41          & 34.39          \\
\multicolumn{1}{c|}{}                          & \multicolumn{1}{l|}{SenCLIP-AttPool} & \textbf{34.61} & \textbf{34.88} & \textbf{34.80} \\ \hline

\multicolumn{1}{c|}{\multirow{6}{*}{ViT-B/32}}  & \multicolumn{1}{l|}{GeoRSCLIP*}               & 41.95 & 37.36 & 32.10          \\
\cline{2-5} 
\multicolumn{1}{c|}{}                          & \multicolumn{1}{l|}{CLIP}                    & 29.80          & 29.50          & 28.37          \\
\multicolumn{1}{c|}{}                          & \multicolumn{1}{l|}{RemoteCLIP}              & 27.17          & 26.87          & 27.76          \\
\multicolumn{1}{c|}{}                          & \multicolumn{1}{l|}{SkyCLIP50}               & 20.16          & 29.87          & 20.21          \\ 
\multicolumn{1}{c|}{}                          & \multicolumn{1}{l|}{SenCLIP-AvgPool}         & \textbf{34.72}          & \textbf{36.78}          & \textbf{37.40} \\
\multicolumn{1}{c|}{}                          & \multicolumn{1}{l|}{SenCLIP-AttPool} & 33.78          & 35.29          & 37.07          \\

\hline
\end{tabular}%
}
\caption{Zero-shot performance comparison on EuroSAT and BigEarthNet using RN50 and ViT-B/32 backbones, highlighting the effect of unified and class-specific prompt strategies. Prompts include a generic format, `\textit{centered satellite photo of \{class\}}', alongside various GPT-3.5-generated aerial and ground view descriptions. *Note: GeoRSCLIP, trained on BigEarthNet with paired text, is considered supervised rather than zero-shot. }
\label{tab:zs_eurosat}
\end{table}
\begin{figure}[htp]
    \centering
    \includegraphics[scale=0.5]{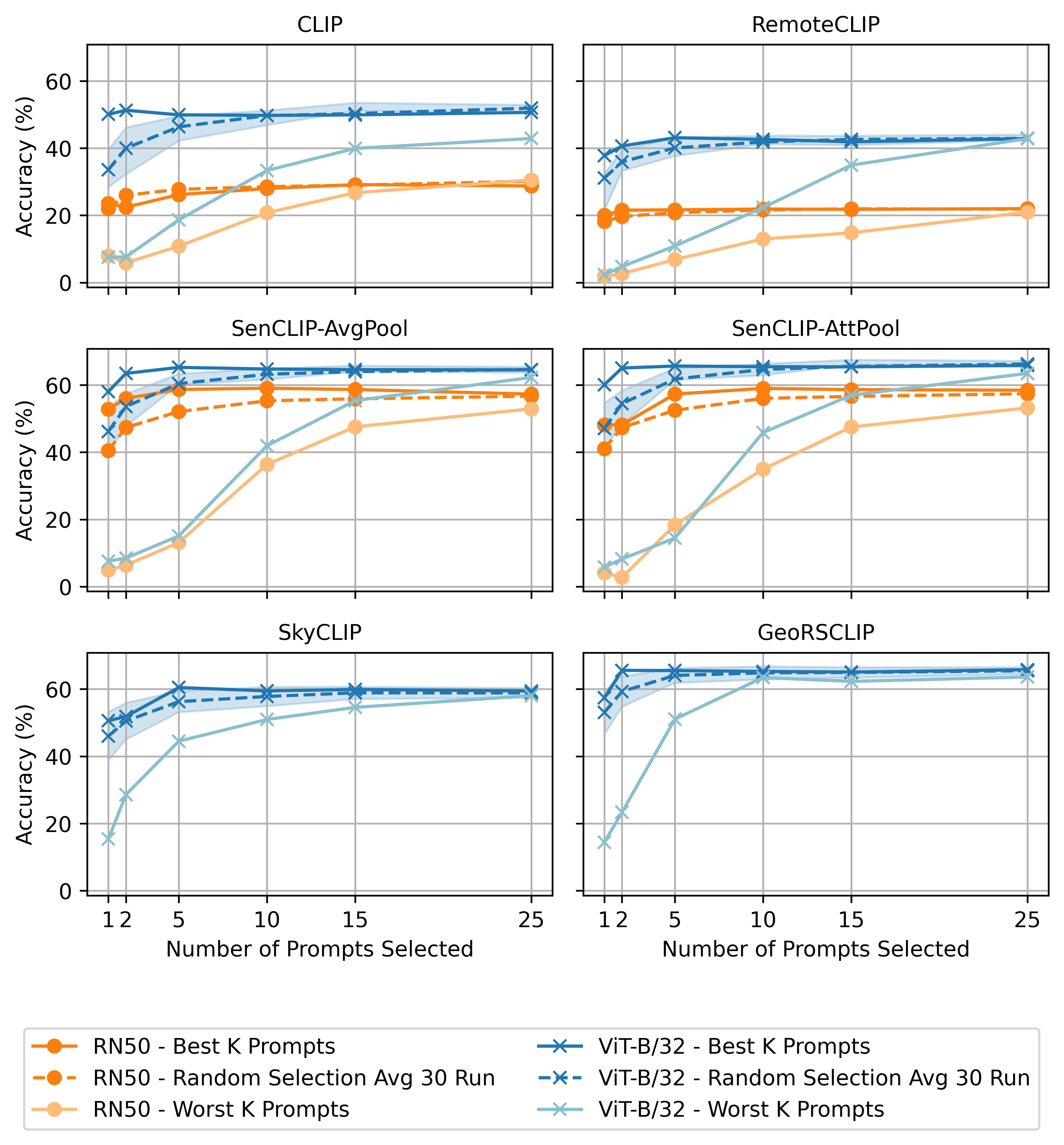} 
  
    \caption{Effect of prompt selection strategies on model (RN50 and ViT-B/32 Backbone) performance on the EuroSAT dataset, varying the number of used prompts. It compares our prompt selection method, which ranks prompts from the most to least descriptive prompts (labelled as Best K and Worst K) for each class, against a random prompt selection baseline. For the random selection baseline we also show the standard deviation over 30 trials. }
    
    \label{fig:ps_zs}
\end{figure}

\noindent\textbf{Zero-shot inference.} Table \ref{tab:zs_eurosat} summarises the zero-shot classification performance of various models using different types of prompts. Models were tested with both generic prompts (e.g., "a centered satellite photo of {class}") and class-specific prompts tailored to aerial and ground-level perspectives. Each class was represented by 50 prompts for each perspective. The comparison includes baseline models like CLIP, RemoteCLIP, SkyCLIP~\cite{wang2024skyscript}, GeoRSCLIP~\cite{zhang2023rs5m}, alongside SenCLIP employing two pooling techniques: AvgPool and AttPool. SenCLIP consistently outperforms all other models, 
demonstrating its superior ability to leverage both aerial and ground-view prompts for enhanced classification accuracy on both dataset with both the aerial and ground-level prompts. 
Specifically, for the ViT-B/32 architecture, SenCLIP improves classification performance by over 10\% for aerial prompts and ground-view prompts compared to CLIP and RemoteCLIP.
While GeoRSCLIP performs competitively with generic satellite prompts, similar to those it has been trained on, its performance declines with more descriptive prompts. It achieves a notable accuracy on the BigEarthNet dataset. However, GeoRSCLIP is trained on BigEarthNet with captions, making it a supervised model rather than zero-shot, limiting its comparability to other models in this context. This is also reflected in the fact that GeoRSCLIP underperforms SenCLIP by 5\% in BigEarthNet when using ground-level prompts, which are further away from the generic ones used to train it.
The class-specific ground-view prompts, along with the unified aerial prompts, showcases the SenCLIP's ability to capture relevant visual and semantic information, leading to superior performance compared to CLIP and remote sensing VLMs.

\noindent\textbf{Prompt selection and zero-shot.}
Fig. \ref{fig:ps_zs} illustrates the impact of prompt selection strategies on model performance on the EuroSAT dataset with RN50 and ViT backbones. Particularly for ViT, the best results are obtained with our approach by selecting a few (2 or 5) prompts. 
For the RN50 backbone, prompt selection with CLIP and RemoteCLIP exhibit similar results to random prompt selection due to their lower effectiveness with ground-level prompts, resulting in varying performance between good and random prompts. 
The worst K-prompt selection, with worse-than-random performance when few prompts are selected, highlights the efficacy of the proposed prompt selection strategy.

\noindent\textbf{Zero-shot classifier tuning (LaFTer~\cite{mirza2023lafter}) on top of SenCLIP.}
Table~\ref{tab:lafter} presents the integration of LaFTer with CLIP and SenCLIP models, revealing the influence of text classifier training on performance. The text classifier trained for the LaFTer default setting of 400 epochs. SenCLIP demonstrates superior performance when employing ground-view and aerial-view prompts (50 per class), for both EuroSAT and BigEarthNet. We found that SenCLIP achieves high level performance with fine-tuning for up to 5 epochs, while CLIP requires 20 epochs to achieve the best results without over-fitting. This could be attributed to SenCLIP being optimised specifically for remote sensing tasks, whereas CLIP has more generalised embeddings that require additional training to optimise. 

\begin{table}[!tbp]
\centering
\resizebox{\columnwidth}{!}{%
\begin{tabular}{l|ll|ll}
\hline
\multicolumn{1}{c|}{\multirow{2}{*}{Model/Prompts}} & \multicolumn{2}{c|}{EuroSAT}                             & \multicolumn{2}{c}{BigEarthNet}                         \\ \cline{2-5} 
\multicolumn{1}{c|}{}                               & \multicolumn{1}{c}{Aerial} & \multicolumn{1}{c|}{Ground} & \multicolumn{1}{c}{Aerial} & \multicolumn{1}{c}{Ground} \\ \hline

CLIP &
  59.88 ± 6.08 &
  48.33 ± 5.79 &
  29.69 ± 5.47 &
  29.17 ± 0.93 \\
SenCLIP AvgPool &
  \textbf{77.90 ± 1.22} &
  72.87 ± 1.12 &
  \textbf{34.67 ± 0.71} &
  \textbf{44.55 ± 1.52} \\
SenCLIP AttPool &
  73.58 ± 0.92 &
  \textbf{75.92 ± 2.13} &
  34.20 ± 1.81 &
  42.79 ± 4.14 \\ \hline
\end{tabular}%
}
\caption{EuroSAT evaluation with LaFTer on top of CLIP and SenCLIP with ViT-B/32 backbone. LaFTer text classifier is trained for 400 epochs and fine-tuned for 20 and 5 epochs for CLIP and SenCLIP, respectively. We used aerial and ground-level prompts. Results are averaged over different seeds.}
\label{tab:lafter}
\end{table}

\subsection{Qualitative results on cross-modal retrieval}

\paragraph{\textbf{Qualitative analysis of SenCLIP embeddings using ClipCap~\cite{mokady2021clipcap}}}
\begin{figure*}[!tp]
    \centering
    \includegraphics[scale=0.24]{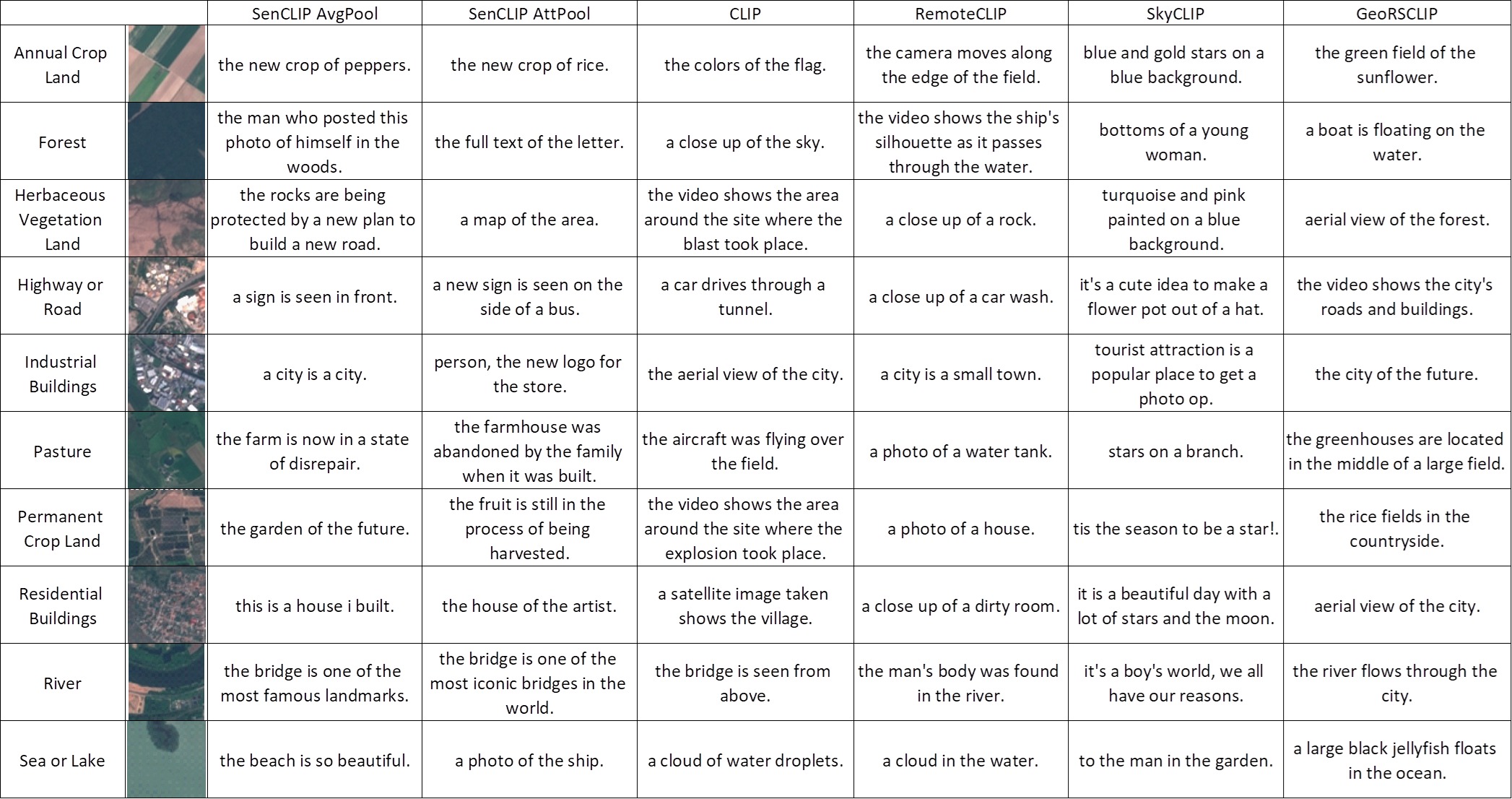} 
    \caption{Image captioning on EuroSAT images using ClipCap~\cite{mokady2021clipcap}}
    \label{fig:clipcap}
\end{figure*}
To further analyse the embeddings learned by SenCLIP, we conducted a qualitative analysis using the ClipCap image caption generator~\cite{mokady2021clipcap}. As shown in Fig. \ref{fig:clipcap}, the captions generated by SenCLIP provide detailed descriptions of the ground-level view. In contrast, captions produced by CLIP and RemoteCLIP predominantly reflect an aerial perspective, often including terms such as \textit{``aerial view"}, \textit{``photo of"} and referencing perspectives from aircraft, video, and camera angles. This distinction underscores SenCLIP's superior ability to capture ground-level details, offering a more nuanced understanding of the scene compared to its counterparts.

\begin{figure*}[!tbp]
  \centering
  \subfloat[\textbf{EuroSAT to LUCAS}]{\includegraphics[ scale =0.32]{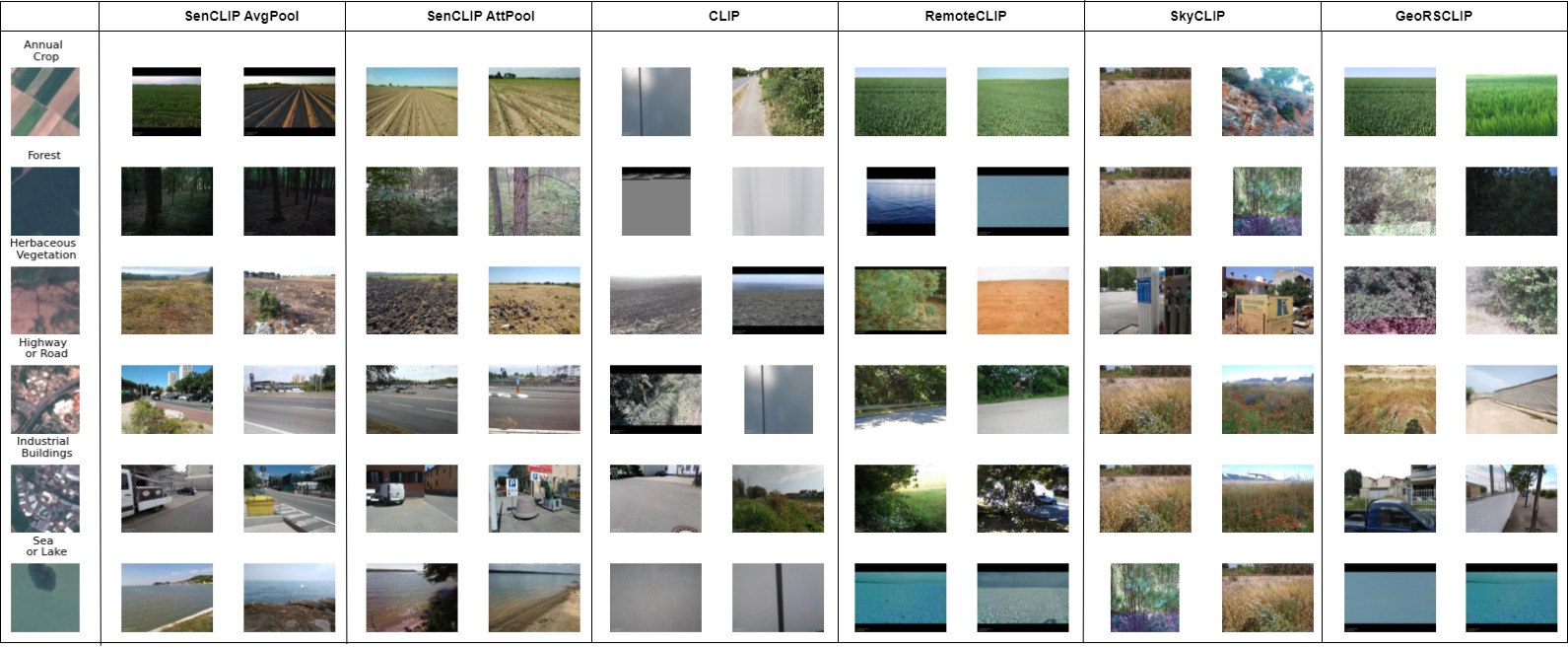}\label{fig:sattogrnd}}
  \vfill
  \subfloat[\textbf{LUCAS To EuroSAT}]{\includegraphics[scale =0.32]{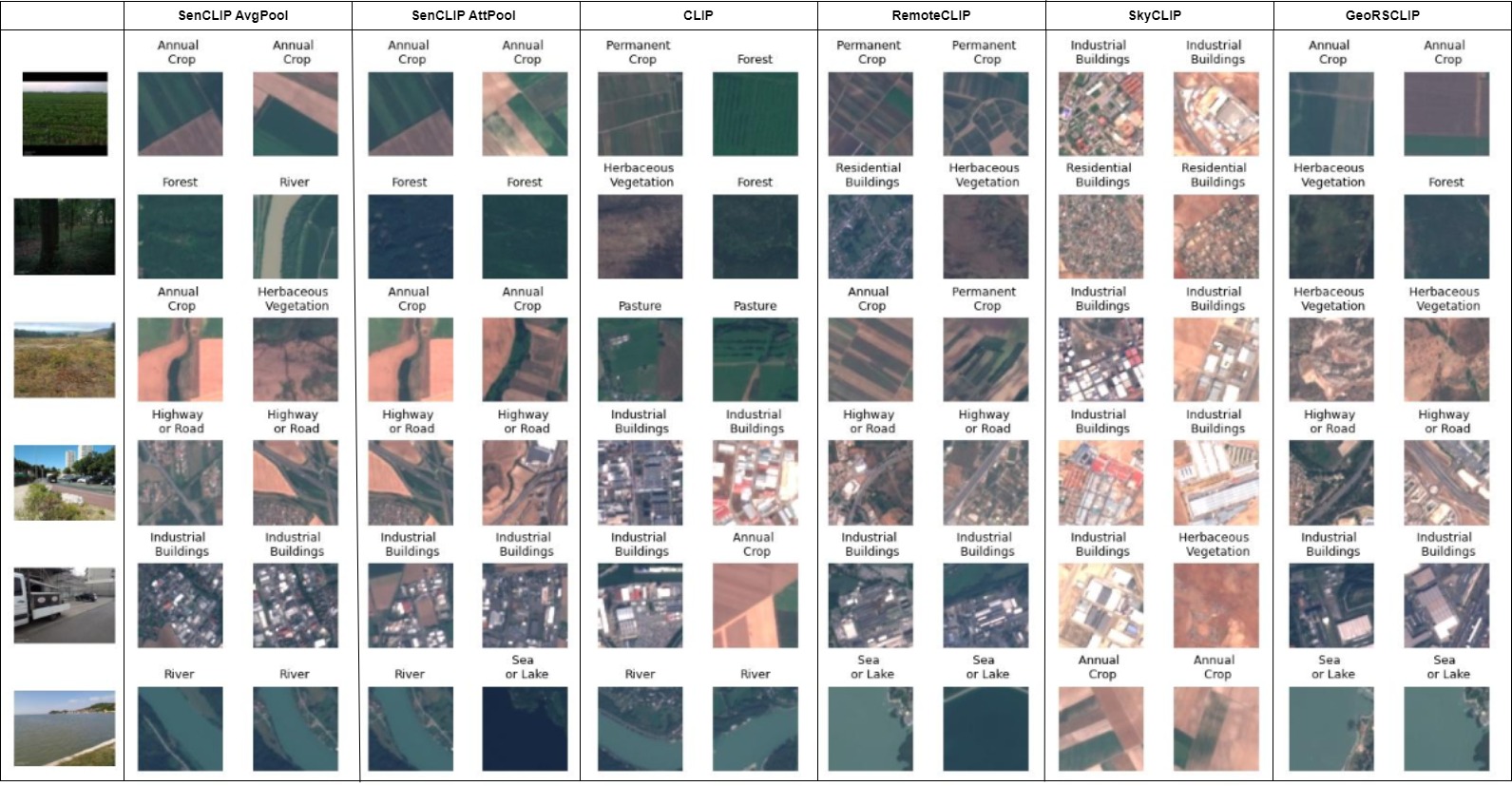}\label{fig:grndtosat}}
  \caption{\textbf{Qualitative image-to-image retrieval}: This analysis demonstrates the qualitative effectiveness of SenCLIP embeddings in both directions. By identifying the top-2 nearest LUCAS embeddings from EuroSAT images, the results indicate that the model successfully learns the fine-grained relationships between ground-level and satellite imagery. Conversely, using ground-level LUCAS images to find the top-2 nearest neighbor satellite images from EuroSAT, the analysis further demonstrates the model's capability to map embeddings bidirectionally, effectively capturing and relating fine-grained details between the two image domains.}
\end{figure*}

\paragraph{\textbf{Image-to-image retrieval - satellite to ground-level.}} We conducted an image-to-image retrieval experiment to identify ground-level images corresponding to a given satellite image. As depicted in Fig. \ref{fig:sattogrnd}, we leveraged EuroSAT images from different classes to identify the two nearest neighbor ground-level LUCAS images. The results demonstrate SenCLIP's strong ability to accurately retrieve ground-level images that align with the given class, significantly outperforming CLIP and RemoteCLIP. The latter models struggled across several classes and frequently selected outlier LUCAS images, highlighting their limitations in establishing robust mappings between satellite and ground-level views. Notably, SenCLIP's fine-grained feature alignment is especially apparent in the \textit{annual crops} class, where the retrieved ground-level images showcase detailed views of plantations.

\paragraph{\textbf{Image-to-image retrieval - ground-level to satellite.}} In our exploration of image-to-image retrieval, we investigated the process of mapping LUCAS ground-level images to EuroSAT satellite images, as shown in Fig. \ref{fig:grndtosat}. For this analysis, we used the top-2 LUCAS images given an example from EuroSAT, as depicted in Fig. \ref{fig:sattogrnd}. The results demonstrate that SenCLIP excels in accurately identifying the correct classes in the majority of cases. However, it is worth noting that CLIP and RemoteCLIP perform better in this scenario than in EuroSAT to LUCAS.

\section{Conclusion}
\label{sec:conc}
This study highlights the benefits of cross-view fine-tuning of CLIP, enabling a remote sensing model to effectively capture ground-level semantic details using medium-resolution Sentinel-2 images. Unlike conventional models, which often struggle with domain-specific terms and predefined class names, our approach demonstrates remarkable flexibility in accommodating diverse prompting styles for zero-shot LULC classification. This flexibility paves the way for the creation of custom LULC maps without requiring any additional training data. The model’s success can be attributed to its comprehensive self-supervised training, which aligns Sentinel-2 representations with CLIP representations of co-located, geotagged, ground-level images from the European Union-wide LUCAS dataset. Furthermore, we introduce an efficient prompt selection method, highlighting the importance of prompt curation. 
Overall, this work combines cross-view training and prompt selection to empower models like SenCLIP, enabling them to surpass the limitations of traditional remote sensing methods. By incorporating ground-level landscape descriptions, SenCLIP sets a new benchmark for zero-shot LULC mapping. \\

\noindent \textbf{Acknowledgements}\\
This research was supported by the ‘Giving Rural Actors Novel Data and Re-Usable Tools to Lead Public Action in Rural Areas’ (GRANULAR)
project, which has received funding from the European Union’s Horizon Europe Research and Innovation Programme under Grant Agreement No.
101061068.

{\small
\bibliographystyle{ieee_fullname}
\bibliography{egbib}
}
\appendix

\input{supplementary}
\end{document}

%% file: supplementary.tex
\appendix{}

\counterwithin{figure}{section}
\counterwithin{table}{section}

\section{Appendices}
\subsection{Dataset Overview}
\label{apx:dataset}
\begin{figure}[bp]
    \centering
    \includegraphics[width=\columnwidth, scale=0.5]{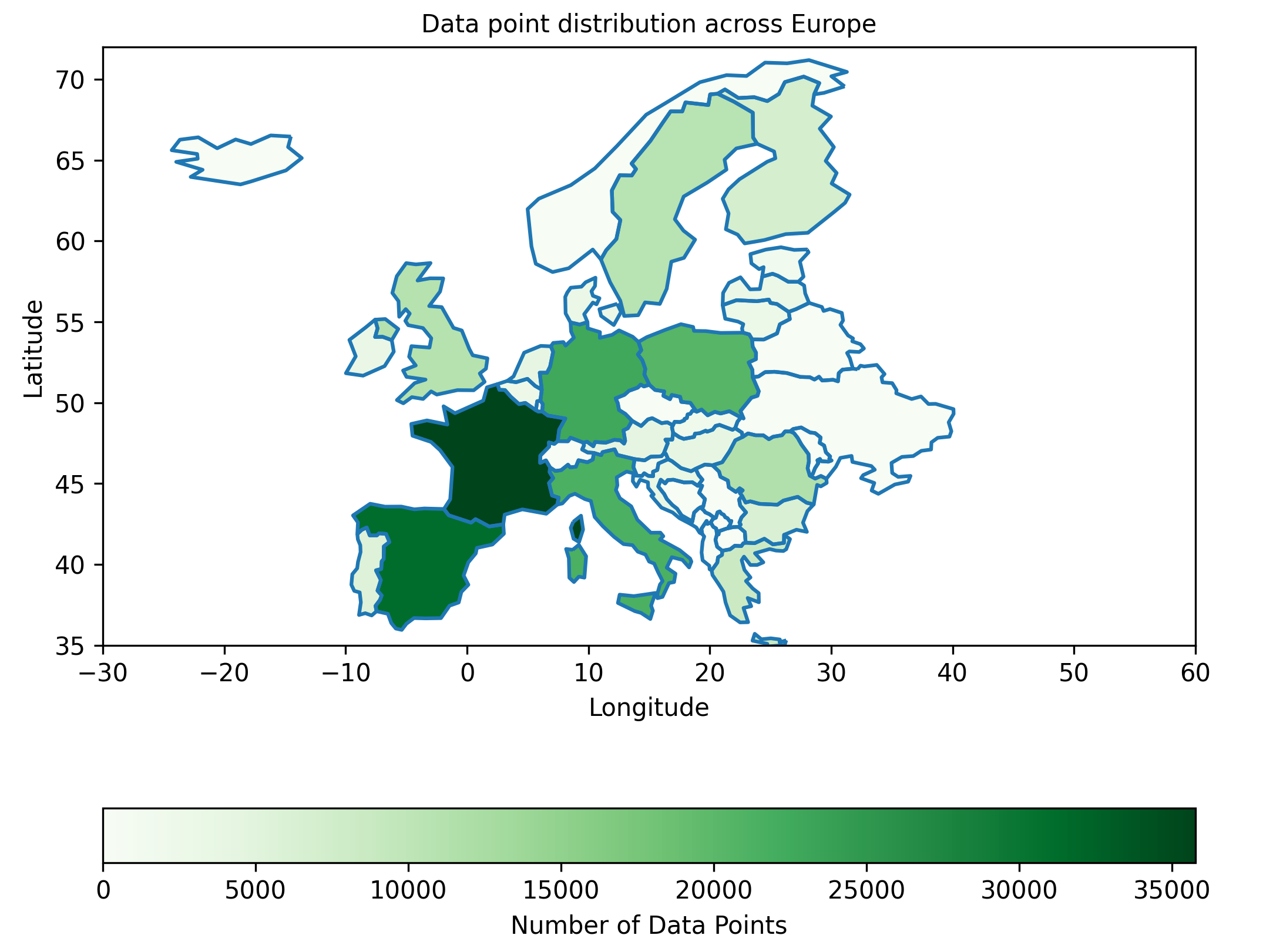}
    \caption{Data Distribution across Europe }
   \label{fig:data_map} 
\end{figure}

\begin{figure*}[tbp]
    \centering
    \includegraphics[width=\linewidth, scale=0.5]{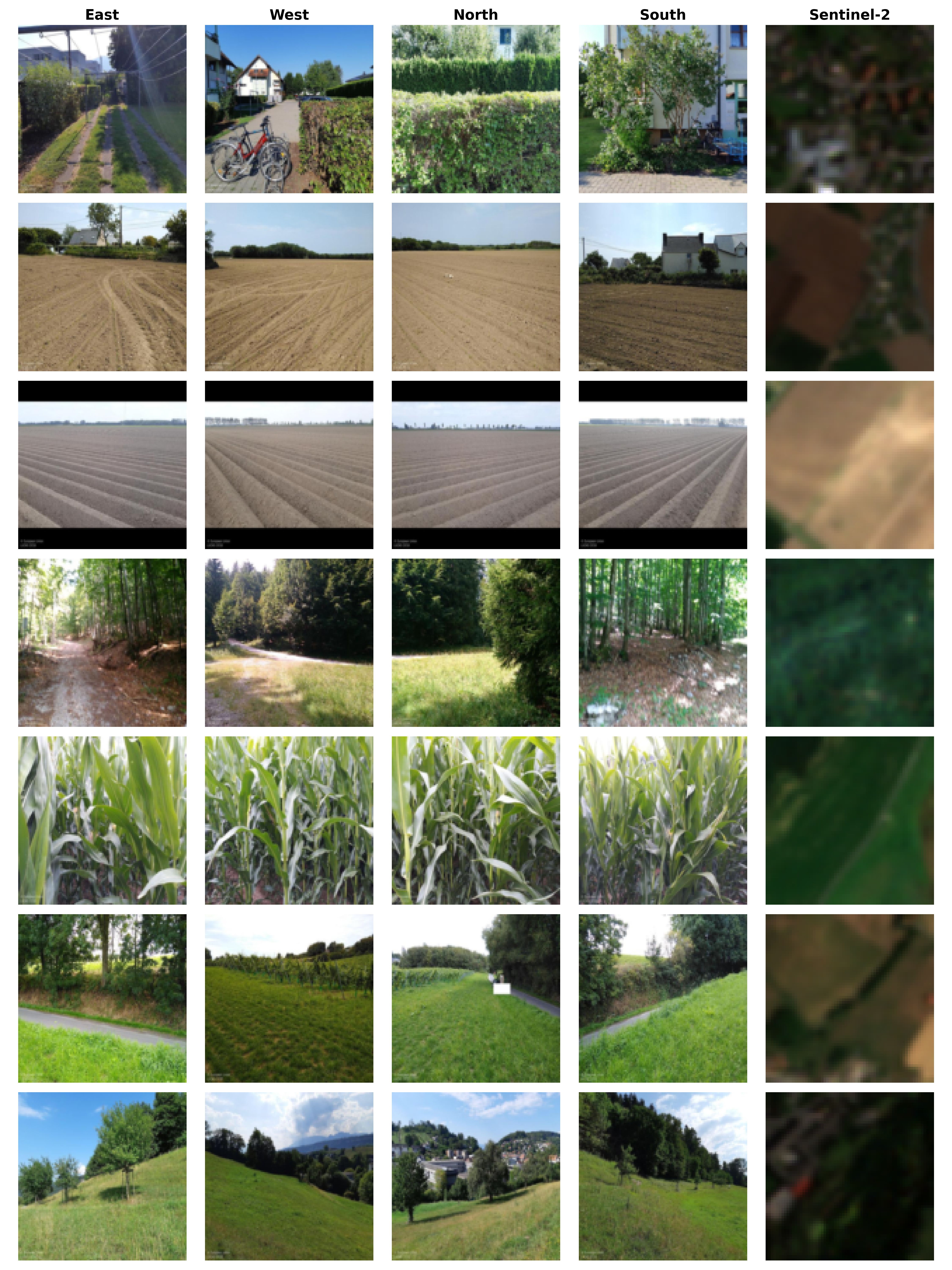}
    \caption{Sentinel-2 collected images from Geo-Tagged LUCAS data points. LUCAS images include four directional views, which are displayed alongside the Sentinel-2 imagery with 10m resolution.}
   \label{fig:data_example} 
\end{figure*}

The LUCAS dataset \cite{d2020harmonised}, encompassing data from 2006, 2009, 2012, 2015, and 2018, includes high-resolution images captured at 1600×1200 pixels. For this work, we focused on the 2018 dataset and downsampled these images to 512×512 pixels using the LANCZOS \cite{duchon1979lanczos} interpolation method. This technique was selected for its superior resampling quality, ensuring the preservation of image detail and clarity. Figure \ref{fig:data_map} illustrates the geographical distribution of geo-tags across Europe, while Figure \ref{fig:data_example} showcases examples of the four directional LUCAS images alongside their corresponding Sentinel-2 images obtained from the Planetary API.

\subsection{Meta Prompts and Prompt Example}
\label{apx:prompt examples}
\begin{figure*}[htbp]
    \centering
    \includegraphics[width=\linewidth, scale=1.1]{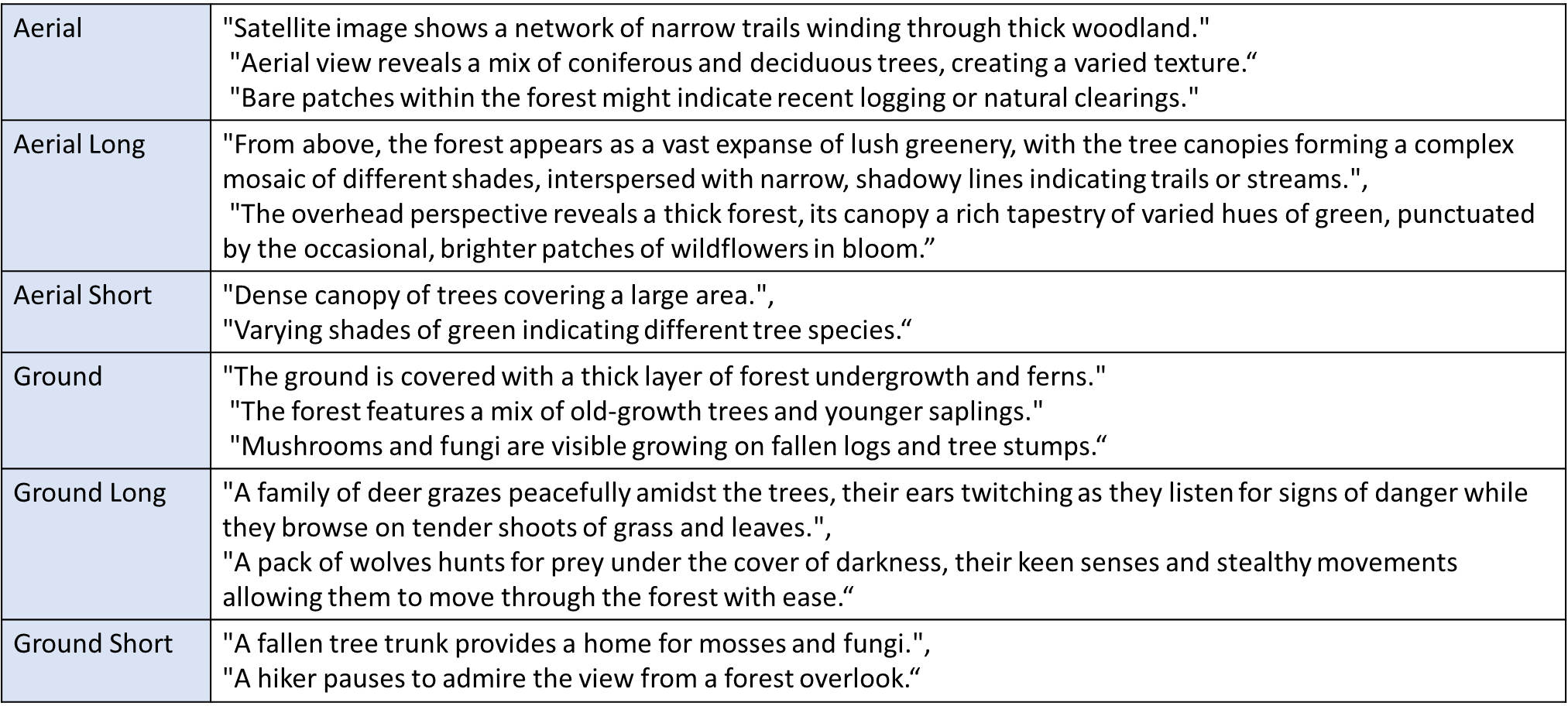}
    \caption{Examples of different style prompts generated for \textit{"Forest}" class by GPT3.5 \cite{ouyang2022training} }
   \label{fig:prompt_llm} 
\end{figure*}

In this work, we generated ground and aerial prompts using the meta-prompt approach described in ``\textit{Meta Prompt for Ground View Prompts}'', and further refined them based on different views, such as ``aerial'' shown in ``\textit{Meta Prompt for Aerial View Prompt}s'', as well as prompt length. An example of the generated prompts for the ``Forest'' class is shown in Fig \ref{fig:prompt_llm}.


\begin{tcolorbox}[colback=gray!10, colframe=black, title=Meta Prompt for Ground View Prompts] \label{gp_metaprompt}
Generate 50 extremely short and diverse sentences that may correspond to factual visual descriptions of photos taken over the land-use/land-cover class `Annual Crop' such that they are as different as possible from all of these other classes:

[`Industrial', `Pasture', `River', `Forest', `Herbaceous Vegetation', `Permanent Crop', `Highway', `Residential', `Sea Lake']

Try to describe visual features or objects that are likely to be visible in such images, even if they are not stereotypical. Make sure they cover as many of all the possible random photos that could be taken over that land-use/land-cover and that they sound as objective as possible, covering different seasons and states of the land-use/land-cover. Make sure to add some examples related to the class for the image visual description. Do not make poetic sentences but more factual.
\end{tcolorbox}

\begin{tcolorbox}[colback=gray!10, colframe=black, title=Meta Prompt for Aerial View Prompts] \label{aerial_metaprompt} 
Generate 50 extremely short and diverse sentences that may correspond to factual visual descriptions of aerial or satellite view over the land-use/land-cover class `Annual Crop'  such that there are as different as possible from all of these other classes:

[`Annual Crop', `Industrial', `Pasture', `River', `Forest', `Herbaceous Vegetation', `Permanent Crop', `Highway', `Residential', `Sea Lake']

Try to describe aerial or satellite  visual features or objects that are likely to be visible in such images, even if they are not stereotypical. Add aerial view context with patterns and use "aerial", "satellite photo" terms.  
Make sure they cover as many of all the possible random photos that could be taken over that land-use/land-cover and that they sound as objective as possible, covering different seasons and states of the land-use/land-cover. Make sure to add some examples related to the class for the image aerial or satellite visual attributes. Do not make poetic sentences but more factual.
\end{tcolorbox}

\subsection{Zeroshot Results based on Length of Prompts}
\label{apx:zs_length}

\begin{table}[htbp]
\centering
 \resizebox{\linewidth}{!}{%
\begin{tabular}{clcccc}
\hline
\multicolumn{2}{c|}{Prompt Templates/Models} &
  Aerial Short &
  \multicolumn{1}{c|}{Aerial Long} &
  Ground Short &
  Ground Long \\ \hline
\multicolumn{6}{c}{EuroSAT} \\ \hline
\multicolumn{1}{c|}{\multirow{4}{*}{RN50}} &
  \multicolumn{1}{l|}{CLIP \cite{radford2021learning}} &
  31.95 &
  \multicolumn{1}{c|}{\textbf{38.42}} &
  28.56 &
  \textbf{31.32} \\
\multicolumn{1}{c|}{} &
  \multicolumn{1}{l|}{RemoteCLIP \cite{liu2023remoteclip}} &
  24.82 &
  \multicolumn{1}{c|}{\textbf{27.80}} &
  \textbf{23.84} &
  20.90 \\
\multicolumn{1}{c|}{} &
  \multicolumn{1}{l|}{SenCLIP-AvgPool} &
  \textbf{57.74} &
  \multicolumn{1}{c|}{\textit{57.02}} &
  \textbf{59.10} &
  55.46 \\
\multicolumn{1}{c|}{} &
  \multicolumn{1}{l|}{SenCLIP-AttPool} &
  \textit{\textbf{58.68}} &
  \multicolumn{1}{c|}{56.34} &
  \textbf{60.12} &
  55.92 \\ \hline
\multicolumn{1}{c|}{\multirow{6}{*}{ViT-B/32}} &
  \multicolumn{1}{l|}{CLIP \cite{radford2021learning}} &
  45.09 &
  \multicolumn{1}{c|}{\textbf{49.61}} &
  40.97 &
  \textbf{41.76} \\
\multicolumn{1}{c|}{} &
  \multicolumn{1}{l|}{RemoteCLIP \cite{liu2023remoteclip}} &
  37.94 &
  \multicolumn{1}{c|}{\textbf{38.85}} &
  \textbf{37.54} &
  35.30 \\
\multicolumn{1}{c|}{} &
  \multicolumn{1}{l|}{SkyCLIP \cite{wang2024skyscript}} &
  \textbf{61.05} &
  \multicolumn{1}{c|}{59.20} &
  53.25 &
  \textbf{54.03} \\
\multicolumn{1}{c|}{} &
  \multicolumn{1}{l|}{GeoRSCLIP \cite{zhang2023rs5m}} &
  62.46 &
  \multicolumn{1}{c|}{\textbf{62.91}} &
  \textbf{60.00} &
  57.88 \\
\multicolumn{1}{c|}{} &
  \multicolumn{1}{l|}{SenCLIP-AvgPool} &
  63.78 &
  \multicolumn{1}{c|}{\textbf{66.89}} &
  \textbf{64.28} &
  58.80 \\
\multicolumn{1}{c|}{} &
  \multicolumn{1}{l|}{SenCLIP-AttPool} &
  \textit{64.10} &
  \multicolumn{1}{c|}{\textit{\textbf{67.54}}} &
  \textbf{63.04} &
  57.82 \\ \hline
\multicolumn{6}{c}{BigEarthNet} \\ \hline
\multicolumn{1}{c|}{\multirow{4}{*}{RN50}} &
  \multicolumn{1}{l|}{CLIP \cite{radford2021learning}} &
  27.60 &
  \multicolumn{1}{c|}{\textbf{30.02}} &
  \textbf{24.41} &
  23.78 \\
\multicolumn{1}{c|}{} &
  \multicolumn{1}{l|}{RemoteCLIP \cite{liu2023remoteclip}} &
  \textbf{32.60} &
  \multicolumn{1}{c|}{32.14} &
  \textbf{31.74} &
  30.66 \\
\multicolumn{1}{c|}{} &
  \multicolumn{1}{l|}{SenCLIP-AvgPool} &
  33.57 &
  \multicolumn{1}{c|}{\textbf{33.60}} &
  30.02 &
  \textbf{32.70} \\
\multicolumn{1}{c|}{} &
  \multicolumn{1}{l|}{SenCLIP-AttPool} &
  \textit{35.18} &
  \multicolumn{1}{c|}{\textit{\textbf{35.89}}} &
  32.74 &
  \textbf{35.09} \\ \hline
\multicolumn{1}{c|}{\multirow{6}{*}{ViT-B/32}} &
  \multicolumn{1}{l|}{CLIP \cite{radford2021learning}} &
  28.58 &
  \multicolumn{1}{c|}{\textbf{34.12}} &
  27.51 &
  \textbf{28.94} \\
\multicolumn{1}{c|}{} &
  \multicolumn{1}{l|}{RemoteCLIP \cite{liu2023remoteclip}} &
  \textbf{32.75} &
  \multicolumn{1}{c|}{28.38} &
  \textbf{29.97} &
  25.44 \\
\multicolumn{1}{c|}{} &
  \multicolumn{1}{l|}{SkyCLIP \cite{wang2024skyscript}} &
  25.77 &
  \multicolumn{1}{c|}{\textbf{28.08}} &
  \textbf{23.43} &
  21.55 \\ \cline{2-6} 
\multicolumn{1}{c|}{} &
  \multicolumn{1}{l|}{GeoRSCLIP* \cite{zhang2023rs5m}} &
  \textit{37.24} &
  \multicolumn{1}{c|}{\textit{\textbf{39.05}}} &
  30.95 &
  \textbf{33.75} \\ \cline{2-6} 
\multicolumn{1}{c|}{} &
  \multicolumn{1}{l|}{SenCLIP-AvgPool} &
  33.08 &
  \multicolumn{1}{c|}{\textbf{35.66}} &
  \textit{34.36} &
  \textit{\textbf{34.57}} \\
\multicolumn{1}{c|}{} &
  \multicolumn{1}{l|}{SenCLIP-AttPool} &
  33.67 &
  \multicolumn{1}{c|}{\textbf{33.76}} &
  33.80 &
  \textbf{33.95} \\ \hline
\end{tabular}%
}
\caption{Zero-shot classification results with RN50 and ViT-B/32 backbones on EuroSAT and BigEarthNet datasets, highlighting the effectiveness of various prompt lengths and types. The comparison includes specific class prompts with short (10-word) and long (50-word) sentence descriptions for aerial and ground views, all generated using GPT-3.5. *Note: GeoRSCLIP \cite{zhang2023rs5m}, trained on BigEarthNet with paired text, is considered supervised rather than zero-shot. \textbf{Bold} indicates the each model's performance on short versus long prompts, while \textit{\textbf{italic}} highlights the overall best-performing model across short and long.}
\label{tab:zs_length}
\end{table}

In addition to generating prompts from aerial and ground perspectives, we further diversified the prompt styles by incorporating varying lengths: short sentences (10 words) and long sentences (50 words), tailored to the specific class under consideration, as illustrated in Fig \ref{fig:prompt_llm}. Table \ref{tab:zs_length} reveals several key insights regarding the impact of prompt length and view-type on zero-shot classification performance. 

\noindent \textbf{Effect of Prompt Length:} Across both the EuroSAT and BigEarthNet datasets, longer prompts (50 words) generally outperform shorter ones (10 words), particularly for aerial views. This trend holds across all models, which show improved accuracy when provided with more detailed prompts. For instance, SenCLIP exhibits notable accuracy improvements with longer prompts on both datasets, especially in aerial views, with the exception of RN50 on EuroSAT.

\noindent \textbf{Ground Views and Prompt Length:} In contrast to aerial views, the effect of prompt length is less pronounced for ground-level images. Models like RemoteCLIP \cite{liu2023remoteclip} and SenCLIP often perform equally well or slightly better with shorter prompts compared to longer ones. This is likely due to the rich visual context inherent in ground-level images, where detailed descriptions in longer prompts may add limited value. For example, in the EuroSAT dataset, SenCLIP-AvgPool and SenCLIP-AttPool show minimal gains from longer prompts in ground views, suggesting that prompt specificity may matter more than length for ground-level imagery.

Moreover, SenCLIP variants consistently outperform other models in both aerial and ground views across both datasets, demonstrating the robustness of its cross-view strategies in leveraging detailed descriptions. While GeoRSCLIP \cite{zhang2023rs5m}, a supervised model, benefits from longer prompts (particularly in BigEarthNet), it is consistently outperformed by SenCLIP in ground-view scenarios. For example, SenCLIP achieves significant performance gains on ground-level imagery within the BigEarthNet dataset.
